\documentclass[11pt]{article}
\usepackage{amsmath,amssymb,amsthm,mathrsfs,stackrel}
\usepackage{array}
\usepackage{graphicx,psfrag,epsf}
\usepackage{multicol}
\usepackage{booktabs}
\usepackage[round]{natbib}
\usepackage{newtxtext}
\usepackage[subscriptcorrection]{newtxmath}
\usepackage{amsfonts}
\usepackage{bm,amsbsy,bbm}
\usepackage{array}
\usepackage{graphicx}
\usepackage{float}
\usepackage{multirow}
\usepackage{threeparttable}
\usepackage{rotating,lscape}
\usepackage{placeins}
\usepackage{algorithm}
\usepackage{algpseudocode}
\usepackage{xcolor}
\usepackage{url}
\usepackage{xr-hyper}
\usepackage[colorlinks,citecolor=blue,linkcolor=blue,urlcolor=blue]{hyperref}
\setcitestyle{authoryear,round}

\newcommand{\J}{\mathcal{J}}

\newcommand{\T}{\mathcal{T}}
\newcommand{\So}{\mathcal{S}}
\def\E{\mathbb{E}}

\def\t{\top}

\def\d{\mathrm{d}}
\def\vp{\varphi}
\def\Var{\text{Var}}
\providecommand{\dint}{\mathrm d}
\providecommand{\cL}{\mathcal L}
\providecommand{\cB}{\mathcal B}
\providecommand{\cV}{\mathcal V}
\providecommand{\cR}{\mathcal R}

\newtheorem{remark}{Remark}
\newtheorem{assumption}{Assumption}

\newtheorem{theorem}{Theorem}

\newtheorem{corollary}{Corollary}
\newtheorem{definition}{Definition}

\begin{document}

\title{Data Fusion for Errors-in-Variables}

\author{
Huali Zhao\\
School of Mathematics and Statistics, Huazhong University of Science and Technology\\[0.4em]
Molei Liu$^{*}$\\
Department of Biostatistics, Peking University Health Science Center;\\
Beijing International Center for Mathematical Research, Peking University\\[0.4em]
Tianying Wang$^{*}$\\
Department of Statistics, Colorado State University
}

\date{}

\maketitle

\begingroup
\renewcommand{\thefootnote}{}
\footnotetext{\textsuperscript{*}Correspondence to: Molei Liu (moleiliu@bjmu.edu.cn) and Tianying Wang (Tianying.Wang@colostate.edu).}
\endgroup

\thispagestyle{empty}
\baselineskip=20pt

\begin{abstract}
We study errors-in-variables problems in which a target study contains only a single error-prone surrogate of an unobserved exposure, while an external source study provides repeated surrogate measurements from a different population. The measurement error distribution is allowed to depend on the observed error-free variables, and the error-free variable distribution itself may differ between studies. We introduce a conditional transportability assumption that enables the use of external repeated measurements under source-target heterogeneity. Together with additional replicate-error conditions, it identifies the target conditional measurement-error distribution. Building on this identification result, we develop a data-fusion estimator for a broad class of target functionals. The estimator combines conditional deconvolution, flexible nuisance estimation, and orthogonal correction that reduces first-order sensitivity to nuisance estimation. For the proposed estimator, we develop a unified spectral theory covering both diffuse-spectrum and finite atomic-spectrum target functionals, derive a general asymptotic expansion, and establish consistency and target-specific convergence-rate bounds. The resulting convergence-rate bounds depend jointly on the spectral properties of the measurement error, the latent exposure, and the target functional. For finite atomic-spectrum targets, we further establish joint Gaussian and bootstrap limits, yielding inference for smooth moment transformations under an additional centering condition. In the reported simulations, Fuse-EIV has small bias for the primary exposure-related coefficient. Applications to the National Health and Nutrition Examination Survey illustrate how accounting for population heterogeneity and error heteroscedasticity can change empirical conclusions. 
\end{abstract}


\pagestyle{plain}

\begingroup
\allowdisplaybreaks
\section{Introduction}
\subsection{Background and Motivation}
Errors-in-variables (EIV) problems arise when a variable of scientific interest is observed only through an error-prone surrogate. Treating the surrogate as the true variable can produce biased estimates and misleading conclusions \citep{loken2017measurement,groenwold2020measurement}. Statistical methods for measurement-error correction have been developed extensively in regression and density estimation since the 1980s \citep{carroll1984binary,stefanski1990deconvolving,carroll2006nonlinear,li2024update}. These methods generally require information beyond a single surrogate, such as repeated measurements, validation data, or sufficiently specific assumptions about the error distribution \citep{carroll2006nonlinear}. Consequently, a study that records only one error-prone measurement per subject and has no validation data generally cannot identify an unrestricted measurement-error distribution.

This limitation is particularly relevant to secondary analyses of large data resources, such as biobanks, cohort studies, and national surveys, whose measurement designs are fixed before a particular scientific question is posed \citep{rutter2023exploring,ross2024leveraging}. Although the study of interest may contain only a single surrogate, repeated measurements may be available in an external study and can provide information about the measurement process. We call the study defining the population and estimand the {\em target} study and the study providing repeated surrogates the {\em source} study. Importantly, the source may represent a different population and, like the target, need not observe the latent variable directly. Our goal is therefore to use source repeated measurements to correct measurement error in the target study while allowing the two populations to differ.

\subsection{Related Literature}
Classical approaches to measurement-error correction use auxiliary information about the measurement process in different ways. SIMEX and related methods correct for measurement error through extrapolation to the error-free case \citep{stefanski1995simulation,kuchenhoff2006general,zhao2025simulation}, regression calibration estimates a calibration relation \citep{kipnis2012regression}, likelihood and Bayesian analyses specify latent-variable and measurement models \citep{schafer1996likelihood,berry2002bayesian}, and deconvolution-based approaches use the error characteristic function \citep{stefanski1990deconvolving,delaigle2008deconvolution}. 
These methods typically rely on measurement-error information available from the study of interest or from an auxiliary sample assumed to represent its error mechanism. This differs from our setting, where the target study contains neither validation data nor repeated surrogates, while the available repeated measurements come from a potentially different source population. Table~\ref{tab:compare_assumption} summarizes how representative EIV methods compare with our setting in terms of nonlinear models, unknown error distributions, heteroscedasticity, and source-target heterogeneity.

Our approach is based on deconvolution because the target of interest is not restricted to a particular regression coefficient or estimating equation. Regression calibration and corrected-score methods organize measurement-error correction around a specified calibration relation or model-specific moment condition \citep{nakamura1990corrected,schennach2004estimation,hall2007semiparametric,kipnis2012regression}. In contrast, deconvolution recovers features of the latent distribution through Fourier inversion and can therefore accommodate a broad class of target functionals, including regression component moments and density functionals \citep{stefanski1990deconvolving,hall2008estimation,butucea2009adaptive}. When the error distribution is unknown, repeated surrogates can identify the required error characteristic function under additional conditions \citep{li1998nonparametric,delaigle2008deconvolution}. Existing deconvolution methods, however, generally learn this information from repeats or auxiliary errors that directly characterize the target measurement process \citep{hou2025nonparametric,delaigle2026nonparametric}; our repeated measurements instead come from a separate source population.

Our two-study setting also relates to data fusion and transfer learning. Existing methods typically transport regression relationships, source outcomes, or estimating-equation information under covariate or label shift \citep{li2022transfer,liu2023augmented,he2024transfusion,yan2024transfer,lee2025doubly}. Our setting differs in two further respects. First, the transported object is the conditional measurement-error distribution, which is not directly observed but must be identified from source repeated measurements. Second, its characteristic function enters the denominator of the target deconvolution operator, so source-side estimation error can be amplified by the inverse problem. 
More directly related to our setting are studies on transporting measurement-error information across populations. \citet{li2026improving} develop a transportable regression-calibration procedure for linear regression using data from a main study and an external validation study. In contrast, our source study may contain only repeated error-prone surrogates without a gold-standard measurement, and our target is not restricted to a linear regression coefficient. Together, these features extend data-fusion ideas to a latent measurement mechanism identified indirectly from repeated surrogates and make population heterogeneity relevant not only to transport but also to the stability and convergence of target estimation.

\begin{table}[!ht]
\centering
\resizebox{\columnwidth}{!}{
\begin{threeparttable}
\caption{Comparison with major classes of EIV methods in the literature.}
\label{tab:compare_assumption}
    \begin{tabular}{llcccc}
    \toprule
    \textbf{Type of methods} & \textbf{Literature} & \textbf{Nonlinear}\tnote{a} & \textbf{Unobserved $U$}\tnote{b} & \textbf{Heteroscedasticity}\tnote{c} & \textbf{Heterogeneity}\tnote{d}\\
    \midrule
    \multirow{3}{*}{\textit{Extrapolation}}
        & \citet{stefanski1995simulation} & $\checkmark$ & $\times$ & $\times$ & $\times$\\
        & \citet{novick2002corrected}    & $\checkmark$ & $\times$ & $\times$ & $\times$\\
        & \citet{kuchenhoff2006general}  & $\checkmark$ & $\times$ & $\times$ & $\times$\\
    \midrule
    \multirow{2}{*}{\textit{Regression calibration}}
        & \citet{kipnis2012regression}   & $\checkmark$ & $\times$ & $\times$ & $\times$\\
        & \citet{li2026improving} & $\times$ & $\times$ & $\times$ & $\checkmark$\\ 
    \midrule
    \multirow{2}{*}{\textit{Likelihood-based}}
        & \citet{schafer1996likelihood}  & $\times$ & $\times$ & $\times$ & $\times$\\
        & \citet{li2019semiparametric}   & $\checkmark$ & $\times$ & $\checkmark$ & $\times$\\ 
    \midrule
    \multirow{2}{*}{\textit{Bayesian-based}}
        & \citet{berry2002bayesian}      & $\checkmark$ & $\times$ & $\times$ & $\times$\\
        & \cite{chen2025total}          & $\checkmark$ & $\checkmark$ & $\times$ & $\times$\\
    \midrule
    \multirow{3}{*}{\textit{Deconvolution kernel}}
         & \citet{delaigle2008deconvolution} & $\checkmark$ & $\checkmark$ & $\times$ & $\times$\\
        & \citet{hou2025nonparametric}   & $\checkmark$ &  $\checkmark$  & $\times$ & $\times$\\
    & \textbf{Proposed method} & $\checkmark$ & $\checkmark$ & $\checkmark$ & $\checkmark$ \\
    \bottomrule
    \end{tabular}
    \begin{tablenotes}
        \item[a] Whether it can handle nonlinear models/functionals.
        \item[b] Whether it accommodates an unknown measurement-error distribution without directly observed individual errors, including through repeated surrogates.
        \item[c] Whether it can accommodate heteroscedastic measurement error.
        \item[d] Whether it can handle heterogeneity between primary and external studies.
    \end{tablenotes}
\end{threeparttable}
}
\end{table}

\subsection{Our Approach and Contributions}
We propose Fuse-EIV, which combines a single-surrogate target study with repeated surrogates from a separate source population. It uses the source repeats to estimate the conditional measurement-error distribution and transports it to the target study. Fuse-EIV then combines this transported distribution with the target surrogate distribution to estimate target functionals through conditional deconvolution. Because the estimated and transported characteristic function introduces a first-order source-side error that can be amplified by deconvolution, Fuse-EIV augments the plug-in estimator with a cross-fitted source-sample correction.

We make three main contributions. First, we make external repeated measurements usable for measurement-error correction under source-target heterogeneity. Directly borrowing an error distribution from an external sample can be invalid when measurement error depends on observed covariates and their distributions differ across studies. We address this by introducing a novel conditional transportability assumption that allows the source and target populations to differ while still identifying the target conditional measurement-error distribution from source repeats.

Second, we develop a cross-fitted, orthogonally corrected conditional-deconvolution estimator for a broad class of target functionals. A key difficulty is that the transported measurement-error characteristic function must be learned from the source data. Because it enters the denominator of the deconvolution operator, its estimation error can be strongly amplified. We therefore derive a source-sample debiasing correction using a density-ratio-weighted representation of the G\^ateaux derivative of the regularized target functional. This correction removes the leading effect of nuisance estimation and provides a general construction for orthogonal corrections when learned nuisance functions enter through a deconvolution operator.

Third, we develop a unified spectral theory for both diffuse-spectrum and finite atomic-spectrum target functionals. The common expansion separates regularization bias, target-sample variation, source-sample variation, and higher-order nuisance error. A distinctive feature of the data-fusion setting is that the source-side contribution to the convergence rate depends jointly on the spectral properties of the measurement error, the latent exposure, and the target functional. This interaction produces different rate regimes across ordinary- and supersmooth configurations for diffuse-spectrum targets, while finite atomic-spectrum targets can attain the parametric rate under fixed-frequency stability conditions.
For the latter class, we further establish joint Gaussian limits and fixed-nuisance bootstrap validity, and derive inference for smooth moment transformations under an additional centering condition.

\section{Problem Setup and Error Identification}\label{sec:problem_setup}
The target study, denoted by $\T$, determines the population and estimand of interest but contains only one error-prone measurement of the latent variable $X$. Its observed sample is $\{W_j,Z_j\}_{j\in\J_{\T}}$, where $W\equiv W^{(1)}$ is the observed surrogate and $Z=(Z_1,\ldots,Z_p)^\top$ is a vector of observed error-free variables. We consider target functionals of the form
\[
    \E_{\T}\{\Psi(X,Z)\},
\]
where $\Psi(x,z)$ is prespecified. This formulation includes regression moments and density functionals, as illustrated in Section~\ref{sec:method}.

The source study, denoted by $\So$, provides two observed surrogates of $X$, with sample $\{W^{(1)}_j,W^{(2)}_j,Z_j\}_{j\in\J_{\So}}$. The latent variable $X$ is unobserved in both studies. Although we use the same symbols $X$ and $Z$ for the corresponding variables in the two populations, their distributions need not coincide across studies. Let $n_{\T}=|\J_{\T}|$ and $n_{\So}=|\J_{\So}|$. We assume that observations are independent and identically distributed within each study and that the two samples are mutually independent. For $\iota\in\{\T,\So\}$ and a random variable $D$, let $f_{\iota,D}$ denote the density of $D$ in study $\iota$, with $f_{\iota,D\mid Z}$ defined analogously. A notation summary is given in Appendix~A1 (Table~A1).

In the target study, we assume the heteroscedastic measurement error model
\begin{equation}\label{eq:error-model}
    W = X + U,\qquad X\perp U\mid Z,\qquad \E_{\T}(U\mid Z)=0,
\end{equation}
where $U$ denotes the target measurement error. Its conditional distribution may vary with $Z$ and is otherwise left unspecified, allowing measurement variability to depend on observed subject characteristics. For example, in nutritional studies, recall error may vary with age, education, or total energy intake. In the source study, we write $W^{(r)}=X+U^{(r)}$ for $r=1,2$ and assume that, conditional on $Z$, $U^{(1)}$ and $U^{(2)}$ are independent with common density $f_{\So,U\mid Z}$. This assumption concerns the replicate errors rather than the observed surrogates $W^{(1)}$ and $W^{(2)}$, which remain dependent through their shared latent variable $X$.

We next show how the target conditional measurement-error distribution is identified from the source repeated measurements. Define the conditional characteristic function of $U\mid Z=z$ in study $\iota\in\{\T,\So\}$ by
\[
    \vp_{\iota,U\mid Z}(t,z)
    := \E_{\iota}(e^{itU}\mid Z=z)
    = \int e^{itu} f_{\iota,U\mid Z}(u\mid z)\d u.
\]

\begin{assumption}[Conditional transportability]\label{assum:shift}
    For every $z$ in the common support of $Z$, assume
    \[
        f_{U\mid Z}(u\mid z)
        :=f_{\T,U\mid Z}(u\mid z)
        =f_{\So,U\mid Z}(u\mid z)
    \]
    for all $u$. Consequently, $\vp_{U\mid Z}(t,z):=\vp_{\T,U\mid Z}(t,z)=\vp_{\So,U\mid Z}(t,z)$ for all $t$ and such $z$.
\end{assumption}
Assumption~\ref{assum:shift} requires only the conditional error distribution given $Z$ to be shared across studies; the marginal distribution of $Z$ and the conditional distribution of $X\mid Z$ may differ. This assumption is natural when the same measurement procedure is used across studies and $Z$ captures systematic variation in measurement error. Together with Eq~\eqref{eq:error-model}, it also implies $\E_{\So}(U^{(r)}\mid Z)=0$ for $r=1,2$.

\begin{figure}[!ht]
    \centering
    \caption{Latent and observed variables in the target and source studies. Dotted circles denote latent variables and solid circles denote observed variables. The equality between the two studies represents conditional transportability of the measurement-error distribution given $Z$ (Assumption~\ref{assum:shift}). For brevity, we write $f_{\iota,U\mid Z}(u\mid z)$ as $f_{\iota}(u\mid z)$ for $\iota\in\{\T,\So\}$.}
    \includegraphics[width=0.65\linewidth]{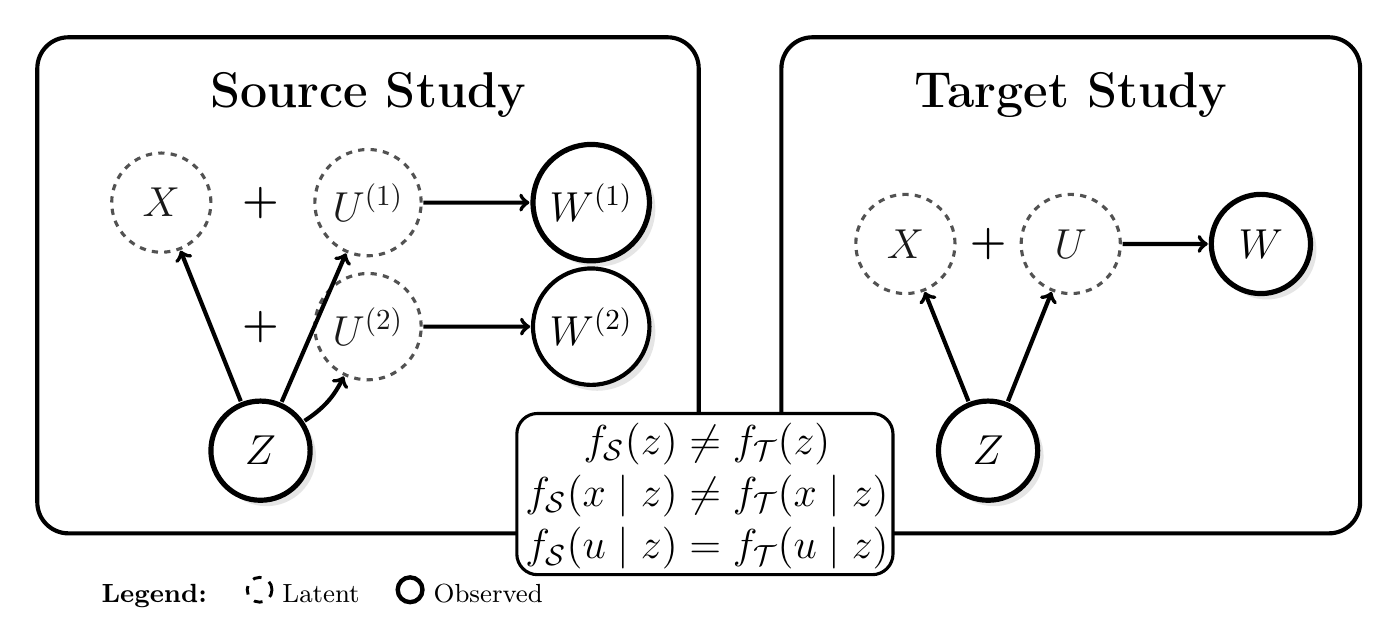}
    \label{fig:two_populations}
\end{figure}

Let $U^d:=W^{(1)}-W^{(2)}=U^{(1)}-U^{(2)}$ denote the observable source replicate difference, and define $\vp_{U^d\mid Z}(t,z):=\vp_{\So,U^d\mid Z}(t,z)$. Conditional independence of the replicate errors then gives
\[
    \vp_{U^d\mid Z}(t,z)
    = \vp_{\So,U\mid Z}(t,z)\vp_{\So,U\mid Z}(-t,z)
    = \big|\vp_{\So,U\mid Z}(t,z)\big|^2.
\]
Thus, the source data identify the modulus of the conditional error characteristic function, but not its phase.

\begin{assumption}[Symmetry and Nonvanishing Characteristic Function]\label{assum:chf_even}
    For each $z$ in the common support of $Z$, assume that the conditional distribution of $U\mid Z=z$ is symmetric about zero and that $\vp_{U\mid Z}(t,z)$ is nonzero for every $t$. Consequently, $\vp_{U\mid Z}(t,z)$ is real, even in $t$, and strictly positive for every $t$ and such $z$.
\end{assumption}

Assumption~\ref{assum:chf_even} holds for common symmetric error models such as Gaussian and Laplace errors \citep{stefanski1995simulation,carroll2006nonlinear}. For brevity, we write $\vp(t,z):=\vp_{U^d\mid Z}(t,z)$. Under the measurement-error model~\eqref{eq:error-model} and Assumptions~\ref{assum:shift} and \ref{assum:chf_even}, for every $t$ and $z$ in the common support,
\begin{align}\label{eq:identify}
    \vp_{\T,U\mid Z}(t,z)
    \equiv \vp_{U\mid Z}(t,z)
    =\sqrt{\vp_{U^d\mid Z}(t,z)}
    \equiv \sqrt{\vp(t,z)}.
\end{align}
Thus, the target conditional distribution of $U\mid Z=z$ is identified on the common support. Under the overlap condition in Assumption~A1, identification holds almost everywhere with respect to the target distribution of $Z$.

\section{The Fuse-EIV Method}\label{sec:method}
In this section, we construct Fuse-EIV for the target quantity
$\E_{\T}\{\Psi(X,Z)\}$ by combining a conditional-deconvolution plug-in
estimator with a source-sample orthogonal correction. We then describe the required nuisance estimators and the extension to regression models.

\subsection{Conditional Deconvolution Plug-in Estimation}\label{subsec:dc}
Let $\vp_K$ be a real, even, bounded, and integrable Fourier multiplier satisfying $\vp_K(0)=1$ and $\operatorname{supp}(\vp_K)\subseteq[-1,1]$, and let
\[
    K(a)=\frac{1}{2\pi}\int e^{-ita}\vp_K(t)\d t
\]
be its inverse Fourier transform. This definition includes an even probability-density kernel but also permits $K$ to be sign-changing. For any finite index set $\mathcal I$, we write $\widetilde{\E}_{\mathcal I}$ for the sample average over $\mathcal I$. In particular, $\widetilde{\E}_{\T}$ and $\widetilde{\E}_{\So}$ average over $\J_{\T}$ and $\J_{\So}$, respectively. For $Z=z$, we define the conditional deconvolution kernel
\begin{equation}\label{eq:K_star}
    K^*_{\lambda}(a,z;\vp)
    := \frac{1}{2\pi}\int e^{-ita}\frac{\vp_K(t)}{\vp_{\T,U\mid Z}(t/\lambda,z)}\d t
    = \frac{1}{2\pi}\int e^{-ita}\frac{\vp_K(t)}{\sqrt{\vp(t/\lambda,z)}}\d t,
\end{equation}
where the second equality follows from the identification identity in Eq~\eqref{eq:identify}, and $\lambda$ is a bandwidth.

\begin{remark}
Under the conditions leading to Eq~\eqref{eq:identify}, $K^*_{\lambda}$ satisfies $\E_{\T}\{K^*_{\lambda}((x-W)/\lambda,Z;\vp)\mid X,Z\}=K((x-X)/\lambda)$. It therefore extends classical homoscedastic deconvolution \citep{stefanski1990deconvolving} to the conditional error model and yields a pointwise density estimator with the oracle kernel-bias order. Section~A2.2 of Appendix~A2 gives the estimator and derivation.
\end{remark}

The population regularized functional associated with Eq~\eqref{eq:K_star} is
\[
    \mathcal T_{\lambda}(\vp)
    :=
    \E_{\T}\left\{\lambda^{-1}\int
    K^*_{\lambda}\big((x-W)/\lambda,Z;\vp\big)\Psi(x,Z)\d x\right\}.
\]
Its difference from $\E_{\T}\{\Psi(X,Z)\}$ is $O(\lambda^2)$ under either standard second-order smoothing conditions or their Fourier-domain analogue; see Section~A2.1 of Appendix~A2 and Proposition~A4. We write $\mathcal T_{\lambda}(\vp)=\E_{\T}\{\Psi(X,Z;\vp)\}$ for the regularized target.

Let $\widehat\vp(t,z)$ estimate $\vp(t,z)=\vp_{U^d\mid Z}(t,z)$ from the source repeats, as described in Section~\ref{subsec:nuisance}. The resulting conditional-deconvolution plug-in estimator is
\begin{align}\label{eq:dc_estimator}
    \widetilde{\E}_{\T}\{\Psi(X,Z;\widehat\vp)\}:= (n_{\T}\lambda)^{-1}\sum_{j\in \J_{\T}}\int K^*_{\lambda}\big((x-W_j)/\lambda,Z_j;\widehat\vp\big) \Psi(x,Z_j) \d x.
\end{align}
Because Eq~\eqref{eq:dc_estimator} substitutes $\widehat\vp$ for $\vp$, its leading error contains a first-order nuisance-estimation term. The deconvolution denominator can amplify this error when the error characteristic function is small, even after regularization. Section~\ref{subsec:fuse-EIV} constructs a correction that reduces this first-order sensitivity.

\subsection{Orthogonal Correction and the Fuse-EIV Estimator}\label{subsec:fuse-EIV}
We construct a Neyman-orthogonal correction and evaluate it by source-sample cross-fitting. The Fuse-EIV estimator subtracts this correction from the plug-in term in Eq~\eqref{eq:dc_estimator}. We view $\mathcal T_{\lambda}(\cdot)$ as a population functional of its characteristic-function argument. Its G\^ateaux derivative in the direction $\widehat\vp-\vp$ gives the leading effect of nuisance estimation. This derivative is defined under the target population, but the target study has no replicate difference with which to evaluate $\widehat\vp(t,z)-\vp(t,z)$. The source study supplies that information through the observable $U^d=W^{(1)}-W^{(2)}$.

To express the target derivative as a source expectation, we define the density ratio and the target conditional characteristic function by
\begin{align}\label{eq:density ratio}
    \omega(z) := \frac{f_{\T,Z}(z)}{f_{\So,Z}(z)},
    \qquad
    \vp_{\T,W\mid Z}(t,z):=\E_{\T}(e^{itW}\mid Z=z).
\end{align}
For an evaluation set $\mathcal I\subseteq\J_{\So}$, the empirical source correction is
{\small
    \begin{align}\label{eq:debias_term}
        \Delta(\omega,\vp_{\T,W\mid Z},\vp; \mathcal I) := -\frac{1}{4\pi\lambda}\widetilde{\E}_{\mathcal I}\left[\omega(Z)\int \int\Psi(x,Z) e^{-itx/\lambda}\vp_{\T,W\mid Z}(t/\lambda,Z)\vp_K(t)\frac{\vp(t/\lambda,Z) - e^{itU^d/\lambda}}{\{\vp(t/\lambda,Z)\}^{3/2}}\d t\ \d x\right].
\end{align}}
The term $e^{itU^d/\lambda}$ is observed in the source sample. Conditionally on the nuisance fits, Eq~\eqref{eq:debias_term} is a feasible source-sample analogue of the leading perturbation of $\mathcal T_{\lambda}(\widehat\vp)$, so subtracting it reduces first-order sensitivity to estimation of $\vp$. Proposition~A2 in Appendix~A2.3 gives the population derivative, source representation, and empirical analogue.

For cross-fitting, we split the source sample into $M$ disjoint folds $\J_1,\ldots,\J_M$ of equal size $n_m=n_{\So}/M$, and let $\J_{-m}=\J_{\So}\setminus\J_m$. For each $m$, we estimate $\omega$ using $\J_{-m}\cup\J_{\T}$ and estimate $\vp$ using $\J_{-m}$. Denote the resulting fits by $\widehat\omega^{[-m]}$ and $\widehat\vp^{[-m]}$. Let $\widehat\vp_{\T,W\mid Z}$ be estimated from the target sample and shared across folds. The fold-specific corrected estimate is
\[
    \widetilde{\E}_{\T}\{\Psi(X,Z;\widehat\vp^{[-m]})\}
    -
    \Delta(\widehat\omega^{[-m]},\widehat\vp_{\T,W\mid Z},\widehat\vp^{[-m]};\J_m).
\]
Cross-fitting separates the source observations used to fit the source nuisances from those used to evaluate the correction. Averaging the fold-specific estimates gives
\begin{equation}\label{eq:proposed_estimator}
    \widehat{\E}_{\T}\{\Psi(X,Z)\} := \frac{1}{M}\sum^M_{m=1}\Big[\widetilde{\E}_{\T}\{\Psi(X,Z;\widehat\vp^{[-m]})\} - \Delta(\widehat\omega^{[-m]},\widehat\vp_{\T,W\mid Z},\widehat\vp^{[-m]};\J_m)\Big].
\end{equation}
Algorithm~\ref{algo:fuse-eiv} summarizes the construction.

\begin{algorithm}[!ht]
\caption{Fuse-EIV}
\label{algo:fuse-eiv}
\begin{algorithmic}[1]
\Require Target data $\{W_j,Z_j\}_{j\in\J_{\T}}$, source data $\{W^{(1)}_j,W^{(2)}_j,Z_j\}_{j\in\J_{\So}}$, target function $\Psi$, number of folds $M$, Fourier multiplier $\vp_K$, its inverse transform $K$, and bandwidth $\lambda$.
\State Split the source sample into $M$ folds $\{\J_m\}_{m=1}^M$ and define $\J_{-m}=\J_{\So}\setminus\J_m$.
\State Estimate $\vp_{\T,W\mid Z}(t,z)$ once from the target sample and use this fit in every fold.
\For{$m=1,\ldots,M$}
    \State Estimate $\omega(z)$ and $\vp(t,z)=\vp_{U^d\mid Z}(t,z)$ using the data in $\J_{-m}\cup\J_{\T}$ and $\J_{-m}$, respectively.
    \State Compute $\widetilde{\E}_{\T}\{\Psi(X,Z;\widehat\vp^{[-m]})\}$.
    \State Compute $\Delta(\widehat\omega^{[-m]},\widehat\vp_{\T,W\mid Z},\widehat\vp^{[-m]};\J_m)$.
\EndFor
\State Average the fold-specific estimators to obtain $\widehat{\E}_{\T}\{\Psi(X,Z)\}$.
\end{algorithmic}
\end{algorithm}

For numerical uncertainty estimation, we use a fixed-nuisance bootstrap procedure. Conditional on the estimated nuisance functions, cross-fitting splits, bandwidth, and numerical integration grid, we resample the target observations and the held-out source observations entering the debiasing correction, while keeping these quantities fixed. For each bootstrap sample, we recompute the estimator in Eq~\eqref{eq:proposed_estimator}. The standard error is estimated by the empirical standard deviation of the bootstrap estimates. For regression applications, the bootstrap component estimates are assembled into the corresponding bootstrap coefficient estimates.

\subsection{Estimation of Nuisance Functions}\label{subsec:nuisance}
The estimator uses three nuisance quantities: the density ratio $\omega(z)$, the target characteristic function $\vp_{\T,W\mid Z}(t,z)$, and the source replicate-difference characteristic function $\vp(t,z)=\vp_{U^d\mid Z}(t,z)$. For density-ratio estimation, let $S$ be the study indicator in the pooled target and source data, with $S=0$ for the target study and $S=1$ for the source. Then
\[
    \omega(Z)
    =
    \frac{f(Z\mid S=0)}{f(Z\mid S=1)}
    =
    \frac{\Pr(S=1)}{\Pr(S=0)}
    \frac{1-\E(S\mid Z)}{\E(S\mid Z)}.
\]
In practice, we estimate $\E(S\mid Z=z)$ from the pooled sample and replace
$\Pr(S=1)/\Pr(S=0)$ by the empirical ratio $n_{\So}/n_{\T}$.

For the target characteristic function, we have
\[
    \vp_{\T,W\mid Z}(t,z) = \E_{\T}\{\cos(tW)\mid Z=z\} + i\E_{\T}\{\sin(tW)\mid Z=z\}.
\]
We approximate the cosine and sine conditional means by
\begin{align}
\begin{split}
    m_1(t,Z)&:=\E_{\T}\{\cos(tW)\mid Z\}=h_1(Z)^\t b(t)+a_1(t,Z),\\
    m_2(t,Z)&:=\E_{\T}\{\sin(tW)\mid Z\}=h_2(Z)^\t b(t)+a_2(t,Z),
\end{split}
\end{align}
where $b(t)=\{b_1(t),\ldots,b_B(t)\}^\top$ is a frequency basis and $a_l(t,Z)$ is the approximation error for $m_l(t,Z)$, $l=1,2$. A convenient specification is $h_l(Z)^\top b(t)=\zeta_l^{0\top}\{b(t)\otimes r(Z)\}$, with a covariate basis $r(Z)$. Spline bases may be used for low- or moderate-dimensional $Z$, while for higher-dimensional $Z$ the coefficients may be estimated by regularized sparse additive regression \citep{ravikumar2009sparse,raskutti2012minimax}. Appendix~A5 gives a concrete lasso construction and sufficient rate conditions. The fitted target nuisance is
\[
    \widehat\vp_{\T,W\mid Z}(t,z)=\widehat h_1(z)^\top b(t)+i\widehat h_2(z)^\top b(t).
\]

For computation, we choose a frequency grid $t_1,\ldots,t_{K_n}$ and form the subject-level pseudo-responses $\cos(t_kW_j)$ and $\sin(t_kW_j)$. We then fit the varying-coefficient regressions to the expanded rows indexed by $(j,k)$. The source nuisance $\vp(t,z)$ is fitted analogously from $U_j^d=W_j^{(1)}-W_j^{(2)}$. Because it is real under Assumption~\ref{assum:chf_even}, only the cosine pseudo-response is required. Section~\ref{subsec:extension} gives the corresponding target fit for moments with an observed multiplier.

\subsection{Application to Regression Models}\label{subsec:extension}
Let $Y$ be a response and $V$ the observed error-free covariates in the target regression model. Under nondifferential measurement error, where $Y\perp W\mid X,V$, one may take $Z=V$. Under differential measurement error, when $Y$ is observed in both studies, one may instead take $Z=(Y,V^\top)^\top$ so that the conditional measurement-error distribution is modeled given the observed variables that may affect it.

The construction extends componentwise to moments $\E_{\T}\{H\psi(X,Z)\}$, where $H$ is observed in the target study and is either included in $Z$ or satisfies $H\perp U\mid X,Z$. Multiply the target summand in Eq~\eqref{eq:dc_estimator} by $H$, replace $\Psi(x,Z)$ with $\psi(x,Z)$, and replace $\vp_{\T,W\mid Z}$ in Eq~\eqref{eq:debias_term} with $\E_{\T}\{He^{itW}\mid Z=z\}$. This conditional Fourier moment can be estimated by applying the varying-coefficient regressions in Section~\ref{subsec:nuisance} to $H\cos(tW)$ and $H\sin(tW)$. Setting $H\equiv1$ recovers the original construction.

For a known mean function (inverse link) $g$ and vector-valued transformation $\xi(X)$, we define $\beta=(\beta_x^\top,\beta_v^\top)^\top$ through
\begin{align}\label{eq:estimating_equation}
    \E_{\T}\left[
    \begin{pmatrix}
        \xi(X)\\
        V
    \end{pmatrix}
    \big\{Y-g(\xi(X)^\t\beta_x+V^\t\beta_v)\big\}
    \right]=0.
\end{align}
For the identity link, let $R=(\xi(X)^\top,V^\top)^\top$. One can estimate the entries of $m=\E_{\T}(RY)$ and $Q=\E_{\T}(RR^\top)$ directly when they involve only $(Y,V)$ and by Fuse-EIV when they involve $\xi(X)$. When $\widehat Q$ is nonsingular, the resulting coefficient estimator is $\widehat\beta=\widehat Q^{-1}\widehat m$. Linear regression with $\xi(X)=X$ and polynomial regression with $\xi(X)=(X,\ldots,X^m)^\top$ use moments such as $\E_{\T}\{\xi(X)Y\}$, $\E_{\T}\{\xi(X)V^\top\}$, and $\E_{\T}\{\xi(X)\xi(X)^\top\}$. 

For a non-identity link and fixed $\beta$, each component of Eq~\eqref{eq:estimating_equation} can be estimated using the target-functional and observed-multiplier constructions above. Specifically, terms involving $Y$ use $H=Y$, while the remaining terms are estimated directly by Fuse-EIV. Let $\widehat U(\beta)$ be the resulting estimating function and $\widehat J(\beta)$ its analytic or numerical Jacobian. Starting from an initial estimator, such as the naive estimator, one can update $\beta^{(r+1)}=\beta^{(r)}-\widehat J(\beta^{(r)})^{-1}\widehat U(\beta^{(r)})$ until convergence. A converged solution of $\widehat U(\beta)=0$ defines a computational extension of Fuse-EIV. Appendix~A6 gives further details and the corresponding algorithm.

\section{Theoretical Properties}\label{sec:theory}
We derive a common orthogonal expansion for diffuse-spectrum and finite atomic-spectrum targets. We make their different spectral requirements explicit through a common operator, specialize the resulting rates, and develop inference for finite atomic-spectrum targets. Throughout this section, let
$
    \phi(u,z):=\vp_{\T,W\mid Z}(u,z).
$
Let \(n=n_{\So}\), assume \(n_{\T}\asymp n\), and take a fixed number of source folds with \(n_m\asymp n\). The supporting assumptions, proofs, and detailed rate and inference results are provided in Appendices~A3-A5.

\subsection{A Unified Spectral Representation}
A diffuse target, such as a pointwise density value, distributes spectral weight over a continuum of frequencies. Polynomial and trigonometric moments instead have Fourier transforms supported at finitely many frequencies, sometimes through derivatives of Dirac masses. The following definition covers both cases.

\begin{definition}[Finite-order spectral target]\label{def:spectral_target}
For each \(z\), let \(\Psi^*(\cdot,z)\) denote the Fourier transform of \(\Psi(\cdot,z)\) in the sense of tempered distributions. We call \(\Psi\) a \emph{finite-order spectral target} if, for some finite integer \(m_\Psi\), one can fix a representation
\begin{equation*}
  \Psi^*(\cdot,z)
  =
  \sum_{\ell=0}^{m_\Psi}D^\ell\mu_{\Psi,\ell,z},
\end{equation*}
where \(D^\ell\) is the \(\ell\)th distributional derivative and each \(\mu_{\Psi,\ell,z}\) is a locally finite complex measure, so that its total variation is finite on every compact frequency interval. Fix any representation satisfying the conditions below for the analysis that follows.
\end{definition}

If \(\Psi^*(u,z)\) is an ordinary locally integrable Fourier transform, take \(m_\Psi=0\) and
\(
  \mu_{\Psi,0,z}(\dint u)=\Psi^*(u,z)\dint u.
\)
If
\(
  \Psi^*=\sum_{\nu}\sum_{\ell}a_{\nu,\ell}(z)\delta_\nu^{(\ell)},
\)
where \(\delta_\nu\) denotes the Dirac measure at frequency \(\nu\) and \(\delta_\nu^{(\ell)}=D^\ell\delta_\nu\), take
\(
  \mu_{\Psi,\ell,z}=\sum_\nu a_{\nu,\ell}(z)\delta_\nu.
\)
Thus the same representation covers diffuse spectra, finite atomic spectra, and mixtures of the two. For derivative-order components (\(\ell\ge1\)), the expansion below uses the finite atomic support condition stated in Appendix~A3.2.

For a Fourier-side multiplier \(h(u,z)\), define 
\begin{equation*}
  \cL_{\Psi,\lambda,z}[h]
  :=
  \frac{1}{2\pi}
  \left\langle
    \Psi^*(\cdot,z),
    \vp_K(\lambda\cdot)h(\cdot,z)
  \right\rangle,
\end{equation*}
where \(\langle\cdot,\cdot\rangle\) denotes the bilinear Fourier-domain pairing, interpreted distributionally when needed; for ordinary functions, \(\langle f,g\rangle:=\int_{\mathbb R}f(u)g(u)\,\d u\). Because \(\vp_K(\lambda u)\) vanishes for \(|u|>1/\lambda\), Definition~\ref{def:spectral_target} gives
\begin{equation*}
  \cL_{\Psi,\lambda,z}[h]
  =
  \frac{1}{2\pi}
  \sum_{\ell=0}^{m_\Psi}(-1)^\ell
  \int_{|u|\le 1/\lambda}
    D_u^\ell\{\vp_K(\lambda u)h(u,z)\}
  \dint\mu_{\Psi,\ell,z}(u).
\end{equation*}
For a diffuse-spectrum target, the operator integrates over a frequency interval. For a finite atomic-spectrum target, it reduces to finitely many evaluations or derivatives at fixed frequencies. The terms \(x^2\), \(\sin x\), and \(\cos x\) are smooth as functions of \(x\), although their Fourier spectra are singular.

\subsection{Unified Asymptotic Expansion}\label{subsec:unified_expansion}
The unified expansion requires overlap, target admissibility and denominator stability, target-adapted nuisance-rate conditions, finite leading variances, and nuisance-bandwidth compatibility; full statements are given in Appendix~A3.2. The factor \(\Gamma_\Psi(\lambda)\) quantifies amplification of first-stage errors by the spectral operator. For the sparse additive estimators in Section~\ref{subsec:nuisance}, under the regularity conditions in Appendix~A5, a sufficient rate for the target-adapted \(L_2\) requirement in Assumption~A3 is
\[
  r_n
  =
  \max\left\{
    a_{\phi,n}
    +\sqrt{\frac{s_{\phi,0}\log d_{\phi,n}}{n_{\T}}},
    a_{\vp,n}
    +\sqrt{\frac{s_{\vp,0}\log d_{\vp,n}}{n_{\So}}},
    a_{\omega,n}
    +\sqrt{\frac{s_{\omega,0}\log d_{\omega,n}}
                       {n_{\T}+n_{\So}}}
  \right\}.
\]
Here, \(a_{\cdot,n}\), \(s_{\cdot,0}\), and \(d_{\cdot,n}\) denote the corresponding approximation errors, sparsity levels, and model dimensions. For an atomic derivative target, the same rate applies to the required fixed-frequency derivatives under the derivative regularity conditions in Appendix~A5, which also gives the full definitions and proof.

The expansion separates four sources of error: regularization bias, target-sample fluctuation, source-sample fluctuation, and a higher-order nuisance remainder. We define the first three components before stating the result. First, define the exact regularization bias
\begin{equation*}
  \cB_\Psi(\lambda)
  :=
  \frac{1}{2\pi}
  \E_{\T}
  \left\langle
    \Psi^*(\cdot,Z),
    \{\vp_K(\lambda\cdot)-1\}
    \vp_{\T,X\mid Z}(\cdot,Z)
  \right\rangle.
\end{equation*}
Under the second-order multiplier condition
\(
    |1-\vp_K(t)|
    \leq C_K\min(t^2,1),
\)
together with the spectral regularity condition in Appendix~A3.4, its absolute value satisfies \(b_\Psi(\lambda):=|\cB_\Psi(\lambda)|=O(\lambda^2)\). Then we define the target and source influence terms and their corresponding standard-deviation scales by
\begin{align*}
  g_{\Psi,\lambda}(W,Z)
  &:=\cL_{\Psi,\lambda,Z}
  \left[e^{iuW}\vp(u,Z)^{-1/2}\right],
  \\
  \ell_{\Psi,\lambda}(U^d,Z)
  &:=\frac12\omega(Z)
  \cL_{\Psi,\lambda,Z}
  \left[
    \phi(u,Z)\vp(u,Z)^{-3/2}
    \{\vp(u,Z)-e^{iuU^d}\}
  \right],
  \\
  \sigma_{\T,\Psi}(\lambda)
  &:=
  [\Var_{\T}\{g_{\Psi,\lambda}(W,Z)\}]^{1/2},
  \qquad
  \sigma_{\So,\Psi}(\lambda)
  :=
  [\Var_{\So}\{\ell_{\Psi,\lambda}(U^d,Z)\}]^{1/2},
\end{align*}
where we use the convention \(\Var(Y)=\E|Y-\E Y|^2\) for a complex-valued influence term. The leading stochastic term combines the target empirical fluctuation with the averaged source correction. The overall estimation-error scale also includes the regularization bias:
\begin{align*}
\cV_{\Psi,n}(\lambda)
&:=
(\widetilde{\E}_{\T}-\E_{\T})
\{g_{\Psi,\lambda}(W,Z)\}
-
\frac{1}{M}\sum_{m=1}^M
\Delta\bigl(
    \omega,\phi,\vp;
    \mathcal J_m
\bigr),
\\
a_{\Psi,n}(\lambda)
&:=
b_\Psi(\lambda)
+n_{\T}^{-1/2}\sigma_{\T,\Psi}(\lambda)
+n^{-1/2}\sigma_{\So,\Psi}(\lambda).
\end{align*}

\begin{theorem}[Unified Fuse-EIV expansion]\label{thm:main}
Under the sampling setup in Section~\ref{sec:problem_setup}, the identification conditions in Assumptions~\ref{assum:shift}-\ref{assum:chf_even}, and Assumptions~A1-A5 in Appendix~A3.2,
\begin{equation*}
  \widehat{\E}_{\T}\{\Psi(X,Z)\}-\E_{\T}\{\Psi(X,Z)\}
  =
  \cB_\Psi(\lambda)
  +\cV_{\Psi,n}(\lambda)
  +\cR_{\Psi,n}(\lambda),
\end{equation*}
where  
\begin{align*}
  \cV_{\Psi,n}(\lambda)
  =
  O_p\{n_{\T}^{-1/2}\sigma_{\T,\Psi}(\lambda)\}
  +O_p\{n^{-1/2}\sigma_{\So,\Psi}(\lambda)\},
  \quad 
  \cR_{\Psi,n}(\lambda)
  =
  O_p\left\{
    \Gamma_\Psi(\lambda)
    (r_n^2+n^{-1/2}r_n)
  \right\}.
\end{align*}
If, in addition,
\(
\Gamma_\Psi(\lambda)\{r_n^2+n^{-1/2}r_n\}
  =o\{a_{\Psi,n}(\lambda)\},
\)
then
\begin{equation*}
  \widehat{\E}_{\T}\{\Psi(X,Z)\}-\E_{\T}\{\Psi(X,Z)\}
  =O_p\{a_{\Psi,n}(\lambda)\}.
\end{equation*}
In particular, the estimator is consistent whenever
\(a_{\Psi,n}(\lambda)\to0\).
\end{theorem}

Theorem~\ref{thm:main} provides a common consistency and convergence-rate result. The expansion also shows how the two studies contribute to estimation: the target sample contributes the empirical fluctuation of the regularized deconvolution functional, whereas the source sample contributes the fluctuation from estimating the conditional error distribution. After the correction, first-stage estimation error enters only through the remainder. The target functional affects all four terms through its spectral operator, regularization bias, operator amplification modulus, and target and source stochastic scales.

A bandwidth sequence \(\lambda=\lambda_n\) is \emph{admissible} when it satisfies the target-adapted denominator-stability condition in Assumption~A4 and the nuisance-bandwidth compatibility condition in Assumption~A6.

\subsection{Rate Specialization under Spectral Smoothness}\label{subsec:rates}
The master rate becomes explicit after separating two features. The decay of the conditional characteristic functions of \(U\mid Z\) and \(X\mid Z\) controls the difficulty of recovering high frequencies. The target spectrum determines whether those frequencies are accumulated over an expanding interval or evaluated only at finitely many fixed points. We use \emph{ordinary smooth} (OS) for polynomial characteristic-function decay and \emph{supersmooth} (SS) for exponential decay. These terms are standard for unconditional deconvolution \citep{fan1991optimal}. Here they refer to conditional spectral envelopes indexed by \(Z\).

To represent the conditional OS/SS regimes uniformly, let \(\gamma_U\) and \(\gamma_X\) satisfy, for all \(u\in\mathbb R\),
\begin{equation*}
    \sup_z|\vp_{U\mid Z}(u,z)|^{-1}
    \lesssim e^{\gamma_U(|u|)},
    \qquad
    \sup_z
    \frac{|\vp_{\T,X\mid Z}(u,z)|}{C_X(z)}
    \lesssim e^{-\gamma_X(|u|)}.
\end{equation*}
Here, \(C_X(z)>0\) is a measurable envelope satisfying \(\E_{\So}\{\omega(Z)^2C_X(Z)^2\}<\infty\). Logarithmic index functions describe OS decay, whereas power-type index functions, possibly with logarithmic corrections, describe SS decay. Appendix~A4.4 states the precise envelope conditions.

\paragraph*{Diffuse-spectrum targets.}
For the explicit rate comparison below, we additionally suppose that the target-side stochastic scale is uniformly bounded, i.e., 
$
    \sigma_{\T,\Psi}(\lambda)=O(1).
$
Appendix~A4.1 gives a sufficient condition for this property. For a diffuse-spectrum target, the operator aggregates spectral amplification across frequencies up to \(1/\lambda\). Suppose its spectral envelope satisfies
\begin{equation*}
    \sup_{z\in\mathcal Z}|\Psi^*(u,z)|
    \lesssim
    (1+|u|)^{-s},
    \qquad s>0.
\end{equation*}
Let \(\rho_{\Psi,\lambda}\) denote the source-side spectral bound defined in Appendix~A4.1. Under the envelope conditions above,
\begin{equation*}
    \sigma_{\So,\Psi}(\lambda)
    \lesssim
    \rho_{\Psi,\lambda}
    \lesssim
    \int_0^{1/\lambda}
    \{\sup_{z\in\mathcal Z}|\Psi^*(u,z)|\}
    \exp\{2\gamma_U(u)-\gamma_X(u)\}
    \dint u.
\end{equation*}
Combining this bound with \(\sigma_{\T,\Psi}(\lambda)=O(1)\) and Theorem~\ref{thm:main} gives, for every admissible bandwidth,
\begin{equation*}
    \widehat{\E}_{\T}\{\Psi(X,Z)\}
    -
    \E_{\T}\{\Psi(X,Z)\}
    =
    O_p\left\{
        b_\Psi(\lambda)
        +
        n^{-1/2}(1+\rho_{\Psi,\lambda})
    \right\}.
\end{equation*}

The four OS/SS regimes follow by substituting the corresponding index functions into this integral bound. The OS indices are
$
    \gamma_U(t)=\theta\log(1+t),
$
and
$
    \gamma_X(t)=\theta_x\log(1+t),
$
whereas the SS indices are
$
    \gamma_U(t)=t^\theta/\alpha-\theta_0\log(1+t),
$
and
$
    \gamma_X(t)=t^{\theta_x}/\alpha_x-\theta_{x,0}\log(1+t).
$
Define
$
    \eta_{\rm OO}
    :=
    2\theta-\theta_x-s+1,
$
and
$
    C
    := 2/\alpha - 1/\alpha_x.
$
Table~\ref{tab::rate_summary_short} summarizes the resulting sufficient upper-rate bounds, with derivations in Appendix~A4.4. The rate bounds reflect the balance between amplification from measurement error and spectral decay of the latent exposure and target. With sufficiently rapid latent spectral decay, faster rates remain possible even under supersmooth errors.

\begin{table}[!ht]
\centering
\caption{Sufficient rate bounds for diffuse-spectrum targets with bounded target-side stochastic scale across the four OS/SS combinations.}
\label{tab::rate_summary_short}
\begin{threeparttable}
\small
\renewcommand{\arraystretch}{1.35}
\begin{tabular}{@{}p{0.11\textwidth}p{0.32\textwidth}p{0.48\textwidth}@{}}
\toprule
\textbf{Regime}\tnote{a} & \textbf{Rate bound} & \textbf{Interpretation}\\
\midrule
(OS, OS)
&
\(
\begin{cases}
O_p\left(n^{-1/(\eta_{\rm OO}+2)}\right),
    & \eta_{\rm OO}>0,\\
O_p(n^{-1/2}\log n),
    & \eta_{\rm OO}=0,\\
O_p(n^{-1/2}),
    & \eta_{\rm OO}<0.
\end{cases}
\)
&
Polynomial deconvolution amplification is offset by latent and target
spectral decay.\\
\midrule
(OS, SS)
&
\(O_p(n^{-1/2})\)
&
Exponential decay of the latent exposure spectrum makes \(\rho_{\Psi,\lambda}=O(1)\), and the parametric rate is attainable whenever an admissible bandwidth also satisfies \(b_\Psi(\lambda)=O(n^{-1/2})\).\\
\midrule
(SS, OS)
&
\(O_p\{(\log n)^{-2/\theta}\}\)
&
A logarithmic bandwidth controls the exponential inverse problem, provided that the resulting bandwidth is admissible.\\
\midrule
(SS, SS)
&
Generally logarithmic, with faster rates in favorable cases
&
If \(\theta_x>\theta\), or if \(\theta_x=\theta\) and \(C<0\), latent spectral decay dominates and \(\rho_{\Psi,\lambda}=O(1)\). When \(\theta_x=\theta\) and \(C=0\), the remaining behavior is determined by the polynomial exponents.\\
\bottomrule
\end{tabular}
\begin{tablenotes}[flushleft]
\footnotesize
\item[a]
The regime is ordered as \((U\mid Z,X\mid Z)\). All displayed rates are conditional on bandwidth admissibility and
\(\sigma_{\T,\Psi}(\lambda)=O(1)\).
\end{tablenotes}
\end{threeparttable}
\end{table}

\paragraph*{Finite atomic-spectrum targets.}
For a finite atomic-spectrum target, only finitely many fixed frequencies matter, so the expanding-frequency difficulty disappears. Suppose that each \(\mu_{\Psi,\ell,z}\) is supported on a common finite frequency set \(\mathcal F_\Psi\), the corresponding coefficients are bounded, and the denominator, multiplier derivatives, and required moments are uniformly stable at the frequencies in
\(\mathcal F_\Psi\). Then
\begin{equation*}
\Gamma_\Psi(\lambda)=O(1),
\qquad
\sigma_{\T,\Psi}(\lambda)=O(1),
\qquad
\sigma_{\So,\Psi}(\lambda)=O(1).
\end{equation*}
Hence, for every admissible bandwidth,
\begin{equation*}
\widehat{\E}_{\T}\{\Psi(X,Z)\}
-
\E_{\T}\{\Psi(X,Z)\}
=
O_p(\lambda^2+n^{-1/2}),
\end{equation*}
provided \(b_\Psi(\lambda)=O(\lambda^2)\). In particular, the parametric rate is attained when \(\lambda^2=O(n^{-1/2})\), together with the required nuisance-rate and fixed-frequency denominator-stability conditions.

This class includes the finite atomic spectra used in the paper:
\[
\begin{gathered}
(x^2)^*=-2\pi\delta_0'', \qquad
(\sin x)^*=\frac{\pi}{i}(\delta_1-\delta_{-1}),\\
(\cos x)^*=\pi(\delta_1+\delta_{-1}), \qquad
(\sin^2x)^*=\pi\delta_0-\frac{\pi}{2}(\delta_2+\delta_{-2}).
\end{gathered}
\]
For \(x^2\), the operator uses a derivative at frequency zero, whereas the trigonometric targets require only function values at finitely many fixed frequencies. Appendix~A4.2 gives the corresponding regularity conditions.

\subsection{Inference for Finite Atomic-Spectrum Targets}\label{subsec:atomic_inference}
We next consider a fixed vector
$
    \boldsymbol\tau
    =
    (\tau_1,\ldots,\tau_K)^\top
$
of finite atomic-spectrum moments, including the observed-multiplier moments described in Section~\ref{subsec:extension}. Directly observed moments may also be included through their empirical averages. Let \(\widehat{\boldsymbol\tau}\) denote the corresponding joint estimator
and \(\boldsymbol\tau_\lambda\) its regularized population counterpart.

\begin{theorem}[Joint normality for finite atomic-spectrum targets]
\label{thm:atomic_inference}
Suppose that \(K\) and \(M\) are fixed, \(n_{\T}/n\to\rho\in(0,\infty)\), \(\lambda=\lambda_n\to0\) is deterministic, and the conditions in Assumption~A7 of Appendix~4.3 hold. Then
\[
    \sqrt n
    (\widehat{\boldsymbol\tau}-\boldsymbol\tau_\lambda)
    \rightarrow
    N_K(0,\Sigma),
\]
where \(\Sigma\) is the limiting covariance matrix given in Appendix~A4.3. The corresponding covariance estimator is consistent, and the fixed-nuisance bootstrap in Section~\ref{subsec:fuse-EIV} satisfies
\[
    \sqrt n
    (\widehat{\boldsymbol\tau}^{*}
    -\widehat{\boldsymbol\tau})
    \rightarrow_*
    N_K(0,\Sigma),
\]
where \(\rightarrow_*\) denotes conditional weak convergence in probability.
\end{theorem}

Theorem~\ref{thm:atomic_inference} is centered at the regularized target \(\boldsymbol\tau_\lambda\). To obtain inference for smooth functions of the unregularized target moments, an additional centering condition is needed.

\begin{corollary}[Inference for smooth moment transformations]\label{cor:atomic_smooth}
Suppose the conditions of Theorem~\ref{thm:atomic_inference} hold and let \(\beta=\beta(\boldsymbol\tau)\) be continuously differentiable in a neighborhood of \(\boldsymbol\tau\). Define
$
    \widehat\beta
    =
    \beta(\widehat{\boldsymbol\tau})
    $
and $
    \beta_\lambda
    =
    \beta(\boldsymbol\tau_\lambda).
$
Then
\[
    \sqrt n
    (\widehat\beta-\beta_\lambda)
    \rightarrow
    N(0,\Omega),\qquad \Omega
    =
    D\beta(\boldsymbol\tau)
    \Sigma
    D\beta(\boldsymbol\tau)^\top,
\]
and the fixed-nuisance bootstrap consistently reproduces the same limiting distribution. The corresponding delta-method covariance estimator is consistent.

For any fixed contrast \(a\) with \(a^\top\Omega a>0\), if
\begin{equation}\label{eq:atomic_contrast_bias}
    \sqrt n\,
    a^\top(\beta_\lambda-\beta)
    \to0,
\end{equation}
then the corresponding studentized statistic is asymptotically standard normal. Hence Wald confidence intervals and two-sided normal-reference \(p\)-values are asymptotically valid for \(a^\top\beta\).
\end{corollary}

When the component regularization biases are \(O(\lambda^2)\), \(\lambda=o(n^{-1/4})\) is a sufficient condition for Eq~\eqref{eq:atomic_contrast_bias}. For fixed-dimensional polynomial or trigonometric regression, the coefficient vector is a smooth function of finitely many such moments; therefore Corollary~\ref{cor:atomic_smooth} applies whenever all required component moments satisfy Theorem~\ref{thm:atomic_inference}.

\section{Simulation Studies}\label{sec:simulation}
We examine whether Fuse-EIV can estimate target quantities accurately when the target study contains one surrogate measurement and the source study contains repeated surrogates, in the presence of heteroscedastic measurement error and covariate shift between the studies. We consider Laplace and Gaussian errors, representing ordinary-smooth and supersmooth regimes, in linear and nonlinear regression settings. The two tasks involve regression moments whose generalized spectra are concentrated at finitely many frequencies.

\subsection{Simulation Designs and Comparisons}\label{subsec:simu_setting}
Both settings use a total sample size of $n_{\So}+n_{\T}=1500$ and covariates $Z=(Z_1,Z_2,Z_3)^\top$, with $Z_1\equiv1$. Study membership depends nonlinearly on $Z$, inducing different covariate distributions in the source and target studies. The target sample contains $\{Y_j,W_j^{(1)},Z_j\}_{j\in\J_{\T}}$, whereas the source sample contains $\{W_j^{(1)},W_j^{(2)},Z_j\}_{j\in\J_{\So}}$. The measurement errors are nondifferential and heteroscedastic, with the repeated source errors generated independently conditional on $Z$. Complete distributional specifications and implementation details are given in Appendix~A7.

Setting I uses the linear response model $Y=X\beta_x+Z^\top\beta_z+\epsilon$, where $\epsilon\sim N(0,0.25)$ and $\beta=(\beta_x,\beta_z^\top)^\top=(1,1,1,1)^\top$. Setting II uses the same construction for $X$, $Z$, study membership, and the heteroscedastic measurement errors, with a setting-specific measurement-error scale, but uses $Y=\beta_0+\sin(X)\beta_x+\epsilon$ with $\beta=(\beta_0,\beta_x)^\top=(1,1)^\top$. Its estimating equation involves $\E_{\T}\{Y\sin(X)\}$, $\E_{\T}\{\sin(X)\}$, and $\E_{\T}\{\sin^2(X)\}$, which are estimated using the observed-multiplier construction in Section~\ref{subsec:extension}. These two settings assess estimation of target regression coefficients through linear and nonlinear component moments.

We compare Fuse-EIV with the \textit{naive estimator}, which replaces $X$ by $W^{(1)}$, and \textit{SIMEX} \citep{stefanski1995simulation}. For SIMEX, a common measurement-error variance is estimated as half the sample variance of the source replicate differences. Gaussian perturbations are added to the original surrogate $W^{(1)}$. In the nonlinear setting, $\sin(W^{(1)})$ is recomputed after each perturbation. Using the selected bandwidth scale $\lambda_c$ (Appendix~A7), we evaluate empirical bias, root mean squared error (RMSE), average estimated standard error, and coverage rate over 200 Monte Carlo replicates. The naive estimator uses model-based standard errors, SIMEX uses jackknife standard errors, and Fuse-EIV uses the fixed-nuisance bootstrap described in Section~\ref{subsec:fuse-EIV}, with 500 bootstrap replications per dataset.

\subsection{Results}
Tables~\ref{tab:simu-os} and~\ref{tab:simu-ss} compare estimation and inference performance across the two regression settings and two measurement-error regimes. Fuse-EIV has the smallest absolute bias for the main $X$-related parameter $\beta_x$ in all four settings. For example, under ordinary-smooth errors, the bias for $\beta_x$ in the linear setting is $0.003$ for Fuse-EIV, compared with $-0.170$ for the naive estimator and $0.057$ for SIMEX, while the corresponding biases in the nonlinear setting are $-0.004$, $-0.120$, and $0.082$. In nonlinear regression, Fuse-EIV also has the smallest RMSE, whereas in linear regression its RMSE is not uniformly smaller than those of the competing methods. For $\beta_x$, the coverage rates of Fuse-EIV range from $0.925$ to $0.950$ across the four settings, while the naive and SIMEX intervals exhibit substantial undercoverage.

Replacing $X$ with its contaminated surrogate produces substantial bias for parameters involving $X$. SIMEX uses a Gaussian working error model with a common variance estimated from the source replicates. This model does not capture the conditional heteroscedasticity and also differs in distributional form under Laplace errors. Residual bias remains for the primary regression parameter under this implementation. For $\beta_{z_1}$ and $\beta_{z_3}$ in the linear setting, all methods have similar bias and RMSE. Here, $Z_1$ is the constant component, $X$ is centered, and $Z_3$ is generated independently of $X$. The corresponding coefficient estimates therefore remain nearly unbiased \citep[Chapter~3]{carroll2006nonlinear}.

\begin{table}[!ht]
\centering
\renewcommand{\arraystretch}{1.5}
\setlength{\tabcolsep}{4pt}
\caption{Simulation results with Laplace (ordinary-smooth) errors.}
\label{tab:simu-os}
\resizebox{\columnwidth}{!}{
\begin{threeparttable}
\begin{tabular}{c ccc ccc ccc ccc}
\toprule
& \multicolumn{3}{c}{\bfseries \normalsize Average Bias}
& \multicolumn{3}{c}{\bfseries \normalsize RMSE}
& \multicolumn{3}{c}{\bfseries \normalsize Standard Error}
& \multicolumn{3}{c}{\bfseries \normalsize Coverage Rate}\\
\cmidrule(r){2-4}
\cmidrule(r){5-7}
\cmidrule(r){8-10}
\cmidrule(r){11-13}
& Fuse-EIV & naive & SIMEX
& Fuse-EIV & naive & SIMEX
& Fuse-EIV & naive & SIMEX
& Fuse-EIV & naive & SIMEX\\
\midrule

\multicolumn{13}{c}{
  \textbf{Linear Regression} ($\lambda_c=0.6$)\tnote{a}
}\\
$\beta_x$
& 0.003 & -0.170 & 0.057 & 0.150 & 0.174 & 0.076 & 0.156 & 0.033 & 0.039 & 0.940 & 0.000 & 0.635 \\
$\beta_{z_1}$
& -0.003 & -0.005 & -0.002 & 0.064 & 0.058 & 0.065 & 0.070 & 0.065 & 0.065 & 0.975 & 0.980 & 0.935 \\
$\beta_{z_2}$
& -0.001 & 0.025 & -0.009 & 0.031 & 0.033 & 0.025 & 0.037 & 0.026 & 0.026 & 0.975 & 0.885 & 0.970 \\
$\beta_{z_3}$
& 0.017 & 0.018 & 0.016 & 0.141 & 0.131 & 0.145 & 0.143 & 0.124 & 0.124 & 0.955 & 0.940 & 0.910 \\
\midrule
\multicolumn{13}{c}{
  \textbf{Nonlinear Regression} ($\lambda_c=0.8$)
}\\
$\beta_x$
& -0.004 & -0.120 & 0.082
& 0.061 & 0.128 & 0.101
& 0.061 & 0.046 & 0.047
& 0.950 & 0.240 & 0.580\\
$\beta_0$
& 0.003 & 0.003 & 0.005
& 0.036 & 0.034 & 0.037
& 0.035 & 0.033 & 0.032
& 0.955 & 0.935 & 0.920\\ 
\bottomrule
\end{tabular}
\begin{tablenotes}
\item[a] $\lambda_c$ denotes the bandwidth scale.
\end{tablenotes}
\end{threeparttable}
}
\end{table}

\begin{table}[!ht]
\centering
\renewcommand{\arraystretch}{1.5}
\setlength{\tabcolsep}{4pt}
\caption{Simulation results with Gaussian (supersmooth) errors.}
\label{tab:simu-ss}
\resizebox{\columnwidth}{!}{
\begin{threeparttable}
\begin{tabular}{c ccc ccc ccc ccc}
\toprule
& \multicolumn{3}{c}{\bfseries \normalsize Average Bias}
& \multicolumn{3}{c}{\bfseries \normalsize RMSE}
& \multicolumn{3}{c}{\bfseries \normalsize Standard Error}
& \multicolumn{3}{c}{\bfseries \normalsize Coverage Rate}\\
\cmidrule(r){2-4}
\cmidrule(r){5-7}
\cmidrule(r){8-10}
\cmidrule(r){11-13}
& Fuse-EIV & naive & SIMEX
& Fuse-EIV & naive & SIMEX
& Fuse-EIV & naive & SIMEX
& Fuse-EIV & naive & SIMEX\\
\midrule

\multicolumn{13}{c}{
  \textbf{Linear Regression} ($\lambda_c=0.7$)\tnote{a}
}\\
$\beta_x$
& 0.000 & -0.107 & 0.053
& 0.144 & 0.111 & 0.067
& 0.191 & 0.032 & 0.036
& 0.940 & 0.085 & 0.670\\
$\beta_{z_1}$
& 0.006 & 0.005 & 0.009
& 0.058 & 0.055 & 0.059
& 0.070 & 0.061 & 0.061
& 0.985 & 0.985 & 0.970\\
$\beta_{z_2}$
& -0.001 & 0.015 & -0.009
& 0.032 & 0.028 & 0.027
& 0.042 & 0.025 & 0.025
& 0.970 & 0.935 & 0.925\\
$\beta_{z_3}$
& -0.014 & -0.012 & -0.018
& 0.124 & 0.113 & 0.122
& 0.140 & 0.116 & 0.116
& 0.985 & 0.960 & 0.950\\

\midrule
\multicolumn{13}{c}{
  \textbf{Nonlinear Regression} ($\lambda_c=0.7$)
}\\
$\beta_x$
& 0.002 & -0.077& 0.048 
& 0.064 & 0.089 & 0.074 
& 0.058 & 0.045 & 0.045 
& 0.925 & 0.580 & 0.755 \\
$\beta_0$
& -0.002 & -0.001&0.000
& 0.032 & 0.030 & 0.032
& 0.034 & 0.032 & 0.032
& 0.970 & 0.975 & 0.950\\
\bottomrule
\end{tabular}
\begin{tablenotes}
\item[a] $\lambda_c$ denotes the bandwidth scale.
\end{tablenotes}
\end{threeparttable}
}
\end{table}

\subsection{Sensitivity Analyses}
To examine whether the main conclusions persist at a larger sample size, Tables~A2 and~A3 of Appendix~A7 repeat the simulations after doubling the total sample size from $1500$ to $3000$. Across the four settings, Fuse-EIV has small bias for the primary parameter $\beta_x$, with absolute bias at most $0.026$ and empirical coverage ranging from $0.940$ to $0.975$, while its RMSE decreases with the larger sample size. In contrast, the naive and SIMEX estimators continue to exhibit bias and substantial undercoverage.

To examine sensitivity to bandwidth selection, Tables~A4 and~A5 of Appendix~A7 consider alternative values of the bandwidth scale $\lambda_c$. Across the reported choices, Fuse-EIV again shows stable performance for $\beta_x$, with absolute bias at most $0.024$, similar RMSE, and empirical coverage ranging from $0.920$ to $0.950$.

\section{Applications to NHANES Data}\label{sec:application}
We apply Fuse-EIV to study the association between body mass index (BMI), denoted by $Y$, and long-term total fat intake using data from the National Health and Nutrition Examination Survey (NHANES). Following a similar transformation used in NHANES analyses \citep{carroll2006nonlinear,hou2025nonparametric}, we define the latent exposure as $X=\log(\text{fat intake}+5)$. Because a 24-hour dietary recall is an error-prone measure of long-term intake, we consider two analyses: a primary cross-cycle application using NHANES 2001--2002 as the target and repeated recalls from NHANES 2015--2016 as the source, and a complementary comparison using repeated target-like recalls from NHANES 2017--2018. In both repeated-recall cycles, the first interview was conducted in person and the second by telephone 3 to 10 days later \citep{nchs2018dietary,nchs2020dietary}.

The repeated recalls in NHANES 2015--2016 and 2017--2018 allow us to examine whether replicate variability changes with the observed covariates. Under the conditional replicate-error model,
$
    \E\{(U^d)^2\mid Z\}
    =
    2\operatorname{Var}(U\mid Z).
$
Table~\ref{tab:Ud2_Z} shows that squared replicate differences are associated with energy, age, and education in the source study and with age, race, education, and physical activity in the target-like study, providing evidence of covariate-dependent replicate variability in both samples. Conditional transportability and the remaining identification conditions are maintained as separate assumptions.

\begin{table}[!ht]
\centering
\begin{threeparttable}
\caption{Regression of squared replicate differences $(U^d)^2$ on observed covariates in the source and target-like studies.}
\label{tab:Ud2_Z}
\begin{tabular}{l ccc ccc}
\toprule
&
\multicolumn{3}{c}{\bfseries Source study, NHANES 2015--2016}
&
\multicolumn{3}{c}{\bfseries Target-like study, NHANES 2017--2018}
\\
\cmidrule(lr){2-4}
\cmidrule(lr){5-7}
Covariate
& Estimate
& Standard Error
& $p$-value
& Estimate
& Standard Error
& $p$-value
\\
\midrule
energy
& 0.003 & 0.001 & 0.005
& 0.001 & 0.001 & 0.236 \\
sex
& -0.040 & 0.021 & 0.060
& -0.038 & 0.025 & 0.120 \\
age
& -0.002 & 0.001 & 0.001
& -0.004 & 0.001 & $<0.001$ \\
race
& -0.026 & 0.022 & 0.220
& -0.062 & 0.025 & 0.012 \\
education
& -0.033 & 0.008 & $<6\times10^{-5}$
& -0.036 & 0.010 & $<0.001$ \\
sleep
& ---\tnote{a} & --- & ---
& 0.004 & 0.007 & 0.578 \\
sport
& --- & --- & ---
& -0.016 & 0.008 & 0.037 \\
\bottomrule
\end{tabular}
\begin{tablenotes}[flushleft]
\footnotesize
\item[a] 
Dashes indicate covariates not included in the corresponding
diagnostic regression.
\end{tablenotes}
\end{threeparttable}
\end{table}

\subsection{Cross-Cycle Application}
We treat NHANES 2001--2002 as the target cycle and use repeated dietary recalls from NHANES 2015--2016 to characterize measurement error. The target study contains 4,506 participants, with BMI, one dietary recall $W^{(1)}$, and adjustment covariates. The source study contains 4,157 participants, with two dietary recalls $W^{(1)}$ and $W^{(2)}$ and the same adjustment covariates. The vector $Z$ includes total energy intake, recorded as the variable ``energy'', sex, age, race, and education.

We compare Fuse-EIV with the naive analysis, which replaces $X$ by $W^{(1)}$, and SIMEX, which uses the source replicate differences under a homoscedastic measurement-error model. All three methods target the same coefficient vector in the 2001--2002 target cycle. Fuse-EIV uses 500 fixed-nuisance bootstrap replications. For each coefficient, we multiply the sample standard deviation of the central 90\% of bootstrap estimates by the normal-theory factor (approximately \(1.27\)), rather than using the all-draw standard deviation in Section~\ref{subsec:fuse-EIV}. This scale rule is also used for the target-replicate reference (Appendix~A8). The naive and SIMEX analyses use ordinary least-squares and jackknife standard errors, respectively. We report two-sided normal-reference Wald $p$-values only for the primary exposure coefficient $\beta_x$. All models include an intercept, omitted from the table, and the energy coefficient is reported per 100 kcal/day.

The estimated association between long-term fat intake and BMI differs across methods (Table~\ref{tab:real-world}). For $\beta_x$, the estimates are $0.753$ for Fuse-EIV, $1.302$ for the naive analysis, and $2.759$ for SIMEX. A similar difference is observed for energy: Fuse-EIV gives a smaller estimate in magnitude, whereas the naive and SIMEX analyses give larger negative estimates. The estimates for sex, age, and race are relatively similar across methods. For $\beta_x$, both Fuse-EIV and SIMEX have larger standard errors than the naive analysis, whereas the standard errors for the remaining covariates are generally similar across methods. For the primary exposure coefficient $\beta_x$, the Fuse-EIV $p$-value is $0.111$, which gives a non-significant result at the $0.05$ level, whereas the naive and SIMEX analyses both give $p<0.001$ and statistically significant results.

\begin{table}[!ht]
\centering
\begin{threeparttable}
\caption{NHANES cross-cycle association analysis.}
\label{tab:real-world}
\begin{tabular}{l rrr}
\toprule
& \multicolumn{3}{c}{\bfseries Estimate (Standard Error)} \\
\cmidrule(l){2-4}
Covariate
& Fuse-EIV & Naive & SIMEX \\
\midrule
$X$
& 0.753 (0.473)
& 1.302 (0.271)
& 2.759 (0.437) \\
\quad $p$-value
& 0.111
& $<0.001$
& $<0.001$ \\
\midrule
energy
& $-0.033$ (0.021)
& $-0.056$ (0.015)
& $-0.118$ (0.021) \\
sex
& 0.705 (0.193)
& 0.694 (0.196)
& 0.667 (0.196) \\
age
& 0.010 (0.005)
& 0.009 (0.005)
& 0.008 (0.005) \\
race
& $-0.779$ (0.217)
& $-0.804$ (0.200)
& $-0.875$ (0.200) \\
education
& $-0.114$ (0.072)
& $-0.125$ (0.075)
& $-0.154$ (0.075) \\
\bottomrule
\end{tabular}
\end{threeparttable}
\end{table}

\subsection{Target-Replicate Comparison}\label{subsec:target-replicate}
Because only one dietary recall is available in the 2001--2002 target cycle, the cross-cycle analysis does not provide an internal replicate-based reference. We therefore consider NHANES 2017--2018, in which two dietary recalls are available in the target-like sample. Fuse-EIV uses only the first recall, while both target recalls are used to construct an internal target-replicate reference.

The analysis includes 4,067 target-like participants from NHANES 2017--2018 and 4,139 source participants from NHANES 2015--2016. The covariate set additionally includes weekday sleep hours (``sleep'') and days of vigorous recreational activity (``sport''). Both Fuse-EIV and the target-replicate reference use $\lambda_c=0.9$; details of the reference estimator and its bootstrap inference are given in Appendix~A8.

Table~\ref{tab::case-study} shows that the estimated long-term fat-intake coefficient is $0.646$ for Fuse-EIV and $0.280$ for the target-replicate reference, compared with $1.371$ for the naive analysis and $2.876$ for SIMEX. For $\beta_x$, the Fuse-EIV standard error is $0.424$, which lies between those of the naive analysis ($0.351$) and SIMEX ($0.570$), and is slightly larger than that of the target-replicate reference ($0.337$). For energy, Fuse-EIV and the reference give estimates close to zero ($-0.009$ and $0.008$), whereas the naive and SIMEX analyses give more negative estimates. Estimates for the remaining covariates are generally similar across methods. Thus, for $X$ and energy, the Fuse-EIV estimates are closer to the target-replicate reference than those from the naive and SIMEX analyses. For the primary exposure coefficient $\beta_x$, Fuse-EIV and the target-replicate reference give $p$-values of $0.128$ and $0.406$, respectively, and thus yield the same non-significant conclusion at the $0.05$ level. In contrast, the naive and SIMEX analyses both give $p<0.001$, which indicates statistically significant associations.

\begin{table}[!ht]
\centering
\begin{threeparttable}
\caption{NHANES target-replicate comparison with an internal reference.}
\label{tab::case-study}
\begin{tabular}{l rrrr}
\toprule
& \multicolumn{4}{c}{\bfseries Estimate (Standard Error)} \\
\cmidrule(lr){2-5}
Covariate
& Fuse-EIV & Target ref. & Naive & SIMEX \\
\midrule
$X$
& 0.646 (0.424)
& 0.280 (0.337)
& 1.371 (0.351)
& 2.876 (0.570) \\
\quad $p$-value
& 0.128
& 0.406
& $<0.001$
& $<0.001$ \\
\midrule
energy
& $-0.009$ (0.027)
& 0.008 (0.023)
& $-0.042$ (0.020)
& $-0.111$ (0.029) \\
sex
& 1.237 (0.256)
& 1.248 (0.248)
& 1.214 (0.243)
& 1.165 (0.243) \\
age
& $-0.010$ (0.007)
& $-0.009$ (0.007)
& $-0.011$ (0.007)
& $-0.012$ (0.007) \\
race
& 0.173 (0.258)
& 0.198 (0.249)
& 0.123 (0.245)
& 0.017 (0.247) \\
education
& $-0.234$ (0.094)
& $-0.226$ (0.085)
& $-0.249$ (0.100)
& $-0.279$ (0.100) \\
sleep
& $-0.271$ (0.080)
& $-0.270$ (0.077)
& $-0.273$ (0.071)
& $-0.277$ (0.070) \\
sport
& $-0.407$ (0.071)
& $-0.406$ (0.071)
& $-0.410$ (0.075)
& $-0.416$ (0.075) \\
\bottomrule
\end{tabular}
\end{threeparttable}
\end{table}

\section{Discussion}\label{sec:discussion}
Fuse-EIV estimates target-population functionals by combining the target surrogate distribution with conditional measurement-error information from source repeated measurements. The conditional measurement-error distribution is shared across studies, while the covariate distribution and the conditional distribution of the latent exposure may differ. The target functional then determines the frequencies required for deconvolution. Diffuse-spectrum targets require control over an expanding frequency range as the bandwidth decreases, whereas the finite atomic examples use only fixed frequencies and derivatives.

Across the reported regression settings, Fuse-EIV has small bias for the primary exposure-related coefficient, whereas the naive and SIMEX analyses have substantial bias in several settings. In NHANES, the estimated long-term fat-intake association differs materially across the three measurement-error treatments, and the target-replicate comparison uses a second error-prone recall to construct an internal reference.

Fuse-EIV could be extended to asymmetric measurement error under suitable identification conditions \citep{comte2015density,adusumilli2020inference}. For multivariate latent exposures, we could investigate whether distributional regression methods such as engression \citep{Shen2025Engression} can replace high-dimensional deconvolution. It would also be useful to extend Fuse-EIV to multiplicative measurement error \citep{zhao2024debiased}. External repeated measurements can supply error information that is missing from the target study while the target data continue to determine the estimand.

\endgroup

\paragraph*{Supplementary material}
It includes a notation summary, technical proofs, detailed convergence-rate and finite atomic-spectrum inference results, a computational generalized regression extension, and additional simulation and data-analysis results.

\bibliographystyle{abbrvnat}
\bibliography{ref}

\end{document}